\documentclass[letterpaper, 10pt, conference]{ieeeconf}
\IEEEoverridecommandlockouts{}
\usepackage{cite}
\usepackage{amsmath,amssymb,amsfonts}
\usepackage{algorithmic}
\usepackage{graphicx}
\usepackage{textcomp}
\usepackage{xcolor}
\usepackage{multirow}
\usepackage{tikz}
\usepackage{tikzducks}
\usepackage{duckuments}
\usepackage{diagbox}
\usepackage{listings}
\usepackage{multicol}
\usepackage{subcaption}

\def\BibTeX{{\rm B\kern-.05em{\sc i\kern-.025em b}\kern-.08em
    T\kern-.1667em\lower.7ex\hbox{E}\kern-.125emX}}

\begin{document}

% \title{\LARGE \bf ReTRO Localisation: Raman specTroscopy-dependent mobile RObot localization}
\title{\LARGE \bf RaSpectLoc: RAman SPECTroscopy-dependent robot LOCalisation}

\author{Christopher Thirgood$^{1a}$, Oscar Mendez$^{1a}$, Erin Chao Ling$^{1b}$, Jon Storey$^{2}$, Simon Hadfield$^{1a}$% <-this % stops a space
	% \thanks{*This work was supported by the }% <-this % stops a space
	\thanks{$^{1a}$CVSSP, Computer Science and Electronic Engineering,
		University of Surrey, Guildford, Surrey, United Kingdom
			{\tt\small \{c.thirgood, s.hadfield, o.mendez\}@surrey.ac.uk}}%
	\thanks{$^{1b}$Surrey Institute for People-Centered Artificial Intelligence,
		University of Surrey, Guildford Surrey, United Kingdom
			{\tt\small chao.ling@surrey.ac.uk}}%
	\thanks{$^{2}$Industrial 3D Robotics,
		Tonbridge, Kent, UK
			{\tt\small jstorey@i3drobotics.com }}%
}

\maketitle

\begin{abstract}
    This paper presents a new information source for supporting robot localisation: material composition. 
    The proposed method complements the existing visual, structural, and semantic cues utilized in the literature. 
    However, it has a distinct advantage in its ability to differentiate structurally~\cite{b28}, visually~\cite{b32} or categorically~\cite{b1} similar objects
         such as different doors, by using Raman spectrometers.
	Such devices can identify the material of objects it probes through the bonds between the material's molecules. 
    Unlike similar sensors, such as mass spectroscopy, it does so without damaging the material or environment. 
    In addition to introducing the first material-based localisation algorithm, this paper supports the future growth of the field by presenting a gazebo 
        plugin for Raman spectrometers, material sensing demonstrations, as well as the first-ever localisation data-set with benchmarks for material-based localisation. 
    This benchmarking shows that the proposed technique results in a significant improvement over current state-of-the-art localisation techniques, achieving 16\% more accurate 
        localisation than the leading baseline.
    \newline The code and dataset will be released at: https://github.com/ThirgoodC/RaSpectLoc

\end{abstract}
\vspace{-1.5mm}
\section{Introduction}
Mobile robots have historically relied on depth sensors for localisation tasks, more recently, visual (RGB) sensors have become ubiquitous.
However, these traditional RGB approaches face challenges in urban environments where large planar regions of uniform colour often dominate and distinct visual landmarks are lacking. 
This is especially problematic in self-similar spaces such as hotels with identical rooms and rotational symmetric floorplans.

A novel solution is to sense the material composition of the environment, which can help distinguish visually similar objects, such as different doors. 
Additionally, small deviations in material composition, such as impurities in concrete or uneven levels of corrosion, can serve as local landmarks. 
These observations can be made and the material composition identified using sensors such as mass spectrometers and Raman spectrometers.

In this paper, we propose a new localisation approach that leverages the capabilities of Raman spectroscopy. 
Raman probes are commonly used in physics as active sensors to analyze at a range, without damaging the subject matter. 
They work by analyzing molecular interactions and bonds through light scattering.
This results in a Raman spectrum featuring peaks corresponding to specific molecular bond vibrations.

We present an approach that uses spectral responses produced by a Raman spectrometer as its central sensing unit. 
It can work with or without range information and demonstrates superior performance to vision and depth-based methods. 
Furthermore, it effectively handles inaccuracies introduced in raw responses from the sensor.

Fig. 1 demonstrates a visualisation of RaSpectLoc, where spectra are compared against each other within a material map. 

\begin{figure}[t!]
	\centering
	\includegraphics[width=\columnwidth]{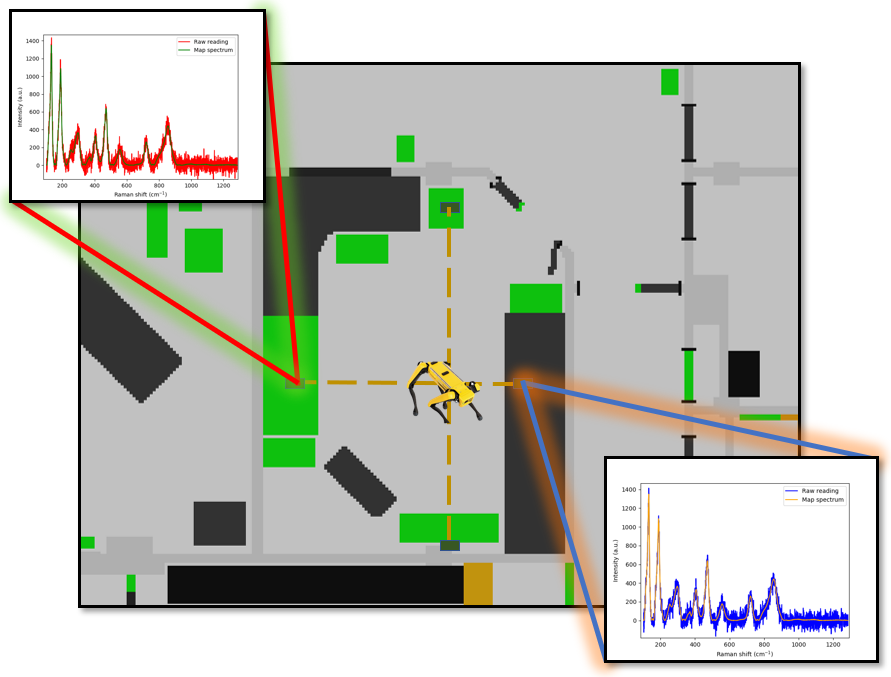}
	\caption{Visualisation of the RaSpectLoc system in action. Spectral readings are compared between the environment and Map along a number of bearing vectors.}
	\label{fi11:vis_demo}
	\vspace{-5mm}
\end{figure} 

In summary, the contributions of this paper are:
\begin{enumerate} 
    \item A novel approach in mobile robotics that utilizes material-based spectroscopic data for localisation. 
	\item A plug-in for the popular simulator, Gazebo, to simulate Raman spectrometers. 
	\item A new dataset containing maps and recordings of material composition from a test environment.
	\item A benchmarked performance evaluation of our approach which provides a far greater average trajectory error compared 
		to current state-of-the-art localisation approaches that utilize RGBD sensors.
\end{enumerate}
	\vspace{-2mm}
\section{Literature Review}
Mobile robot applications have traditionally used scan-matching approaches to localise within a known map. These approaches have historically been dominated by particle filter-based methods, such as AMCL~\cite{b28} and GMapping~\cite{b33} which have been popular solutions due to the integration with the ROS navigation stack~\cite{b7}.  
The technique is range-based (RMCL) and is highly adaptable, adjusting the number of hypotheses generated by the particle filter as confidence in the result improves.
However, AMCL may fail if the map and the sourced environment differ. 
Many `Simultaneous Localisation and Mapping' (SLAM) systems intend to solve this by generating a map at run-time and identifying `loop closures' to solve for drift~\cite{b4, b5} for static environments.
% For unexpected range measurements from the environment, SLAM and RMCL tend to falter as confidence in the correct solution is lost with unlikely range estimates.
For typical dynamic scenes such as industrial work-spaces~\cite{b9, b24} these approaches have attempted to track the pose using an 
    apriori map and secondary short-term map at two different timescales. 
The problem with such a method is that it is too computationally complex for mobile robotics to reliably localize for long periods with accumulated errors.
Our approach aims to resolve this problem via the use of efficient spectral comparison functions which can provide continuous similarity scores.
These are robust to misalignments.

Many RMCL approaches seek to enhance MCL localisation for mobile robots by using 3D point clouds generated from 3D LiDAR systems. 
Maken et al.~\cite{b25} utilized Iterative Closest Point (ICP) with LiDAR sensors for pose estimation and localisation with MCL. 
Although improvements were evident in the results, the accuracy of the covariance estimation of the ICP output poses is dramatically affected by positioning accuracy. 
In recent deployments, machine learning models have become common in localisation systems that utilize point clouds to improve accuracy and reduce covariance estimation.
For example, SegMap proposed by Dube et al.~\cite{b12} uses a Convolutional Neural Network (CNN) to identify and segment geometric primitives in point clouds. 
However, the use of local and global descriptors leads to slow performance due to the expensive descriptor similarity networks. 
As a solution, SegSemMap~\cite{b11} proposed by Cramariuc et al. enriches point clouds with CNN segmentation overlays from RGB cameras and adds them to the map.
These approaches still require a compact and computationally efficient representation of the objects for segmentation, providing a high representation strength.
Although SegMap and SemSegMap do not use an MCL approach, many segmentation-based MCL approaches have been used in combination with geometry and range, like Hendrikx et al.~\cite{b21}. 
Another example is 3D geometry in BIM models which can be used to estimate the camera pose. 
The models are then segmented to gain local information about the area using range readings from a 2D range-based sensor.
Mendez et al.~proposed SeDAR~\cite{b1} which aims to simplify these models with a more robust approach to RMCL by removing the depth readings altogether. 
With a semantic understanding of the environment through a simple CNN, the weights of the particles are adjusted based on the likelihood of particle observations in the segmented map.
In contrast to the discrete semantic label comparisons used in SeDAR~\cite{b1}, our proposed material comparison offers a much finer level of matching granularity. 
Furthermore, computing likelihoods directly from the Raman spectra in the environment without visual classification networks significantly reduces computational costs. 

There have been attempts to extend the applicability of AMCL to non-visual sensing modalities. 
One such example is seen with haptic sensors, which can provide touch-based sensing of the environment~\cite{b39}. 
Buchanan et al.~\cite{b13} have proposed to use data segments from the surroundings with a neural network to update MCL from a map of different terrains. 
However, this approach requires extensive data on the environment's terrain and prolonged pre-training, which can be challenging. 
Another unconventional sensing approach was proposed by Serrano et al.~\cite{b14}, who used the Wi-Fi signal strength of mobile devices for MCL. 
Although this approach is dependent on the Wi-Fi signal data strength, it can be unreliable in complex indoor environments where signals are absorbed through many walls before being sensed by the receiver. 
Visible Light Positioning (VLP) sensors have also been used, with W. Guan et al.~\cite{b15} being able to identify objects in a room using LED patterns from lights and other devices. 
Despite these unconventional sensing methods, they still face challenges with even minor changes in the environment, leading to significant inaccuracies in the MCL algorithm. 
In contrast, Raman Spectrometers have the advantage of treating every part of the environment as an identifying landmark due to its material fingerprint sensing. 

Raman spectrometers are remote scanning instruments often used to identify materials or substances.
Noise in the sensor comes in three forms, shot, dark and read noise which couples to the Poisson distribution of the data, background noise and electronics of the device, respectively~\cite{b26}. 
Such noise can be overwhelmed by gaussian noise from the device in the process of scanning with modern probes.
To identify or recognise similar Raman Spectra, most approaches use a large database with neural networks producing long training and inference times~\cite{b19,b37}.
In contrast, our approach simplifies and extends this by using the likelihood between stored spectra in the map and the latest spectra read from the environment.
De-noising methods such as those proposed by Lussier et al. and Horgan et al.~\cite{b16, b19} show the benefits of using deep-learning models to remove noise, but can also be done via other methods suggested by Zhao et al.~\cite{b22}. 
Analytical likelihood similarity algorithms have been proposed to identify the type of material from the peaks of the spectra as proposed by Foster et al.~\cite{b17}.
We propose alternative Raman spectral similarity functions in RaSpectLoc by providing new algorithms while still delivering a system with low computational complexity.
The most effective similarity functions for Raman spectra necessitate peak classification through algorithmic convolutions~\cite{b20}.
This is infeasible for MCL approaches as they tend to take too long to process each hypothesis. 

The main focus of Raman spectrometry in the literature is remote sensing of chemicals using mobile robots~\cite{b18}, whereas we go further and apply this research for a robot localisation approach.
This paper proposes a novel approach, utilizing innovative localisation techniques and algorithmic similarity functions for mobile robots equipped with underutilized Raman Spectral sensors.
\vspace{-1.8mm}
\section{Problem definition}
Monte-Carlo Localisation by Dellaert et al.~\cite{b28} provides a framework which performs matching to generate pose hypotheses in a map.
The process is as follows: 
\begin{enumerate}
    \item Particles are sampled, either uniformly around a space or in a Gaussian around a pose hypothesis
    \item Particles are propagated through a motion model, typically the robot's odometry with added Gaussian noise.
    \item Each particle is given a weight based on the accuracy of its observations against the map, typically matching a range and bearing scan line to an existing map
    \item A re-sampling step is performed, proportional to the particle's likelihood, before the process repeats.
\end{enumerate}
The current pose $x_t \in SE(2)$, of the robot, can be estimated as a set of possible samples: $\mathbb{S}_t = \{s_t^{i}; i = 1 \dots N\}$,
    given the wheel odometry measurements: $\mathbb{U}_t = \{u_j; j = 1 \dots t\}$ and depth sensor measurements: $\mathbb{Z}_t = \{z_j; j = 1 \dots t\}$ and a 2D map $\mathbb{V}$.
If all previous odometry and depth measurements are equally weighted, the posterior probability $P(s_t^{i} | \mathbb{U}_t, \mathbb{Z}_t, \mathbb{V})$ can be decomposed into an online sequential process:
\begin{equation}
    P(s^i_t | \mathbb{U}_t, \mathbb{Z}_t, \mathbb{V}) = P(s^i_t | u_t, s^i_{t-1}) P( \mathbb{Z}_{t-1} | \mathbb{U}_{t-1}, s^i_{t-1}, \mathbb{V})
\end{equation}
The motion model in the localisation process is determined by the odometry measurements received from the robot, represented by $u_t$. 
This information is used to ``shift'' the particles, assigning a likelihood based on the probability of the final position given the measured odometry. 
The particles are propagated based on $u_t$, with Gaussian noise added to model sensor noises, as follows:
\begin{equation}
    P\left(s_{t}^{i'} | u_{t}, s_{t-1}^{i} \right) \sim \mathcal{N} \left(u_{t} + s_{t-1}^{i}, \Upsilon_{t} \right) \label{eq:2}
\end{equation}
where $\Upsilon_{t}$ is the covariance of the odometry and $\mathcal{N}$ is a normal distribution across the dimensions of SE(2).
    
The sensor model $P(\mathbb{Z}_{t-1} | \mathbb{U}_{t-1}, s^i_{t-1}, \mathbb{V})$ measures how well the range scan fits the pose hypothesis.
The probability of each range-scan $\left( r_{t} \right)$ is estimated under the assumption all scans are independent. 
Two sensor models which are commonly used with AMCL are the ``Beam'' and ``Likelihood-field'' models.
The Beam model is a raycasting operation where a ray is cast starting from hypothesis $s^i_t$ along the bearing $ \theta^{k}_t$ and terminates when an occupied cell is reached.
The likelihood is estimated as:
\begin{equation}
    P_{R}\left(\mathbb{Z}_{t}^{k} | s_{t}^{i'} \right) =  \exp \left(  {\frac{-\left( r_t^{k} - r_t^{k\ast} \right)^2}{2\sigma _o ^{2}}} \right)
\end{equation}
where $\sigma$ is the variance of the sensor measurement noise and $r_t = r_t^k \forall k=0...N$.
The likelihood field model uses a field similar to a chamfer map to quickly estimate the distance to the nearest geometry in the floorplan, eliminating the need for costly raycasting operations. 
The chamfer map is defined as:
\begin{equation}
    C_{i,j} = \min_{k,l}|[i-k, j-l]|,  \mathbb{V}_{k,l} \neq 0
\end{equation}
where $C_{i,j}$ is the cost value of the chamfer map at position $(i,j)$, while $\min$ is a minimum value function and points $(k,l)$ which iterates over the neighbors of $(i,j)$. 
In the likelihood-field model, assuming a Gaussian error distribution, the weight of each particle, $s'$, can be estimated as:
\begin{equation}
    P_{r}(\mathbb{Z}_{t}^{k} | s_{t}^{i'}, \mathbb{V}) = \exp \left( {- \frac{\delta_{o}^{2}}{2\sigma_{o}^{2}}} \right)
\end{equation}
where $\delta_o$ is the value obtained from the distance map and $\sigma_o$ is dictated by the noise characteristics of the sensor.
During runtime, the endpoint of each bearing-range tuple is computed for each pose hypothesis, and its probability is related to the distance stored in the chamfer map. 
The likelihood field model is faster and produces more accurate results compared to the beam model as it is more robust to orientation errors.

RaSpectLoc is similar to AMCL but replaces many of the operations listed above for material-based sensing.
Our method includes adapting a floorplan for Raman spectra, a novel spectral-based raycasting and a particle weight calculation method for each hypothesis. 
	\vspace{-2mm}
\begin{figure}[htp]
	\centering
	\includegraphics[width=0.8\columnwidth, height=6.5cm]{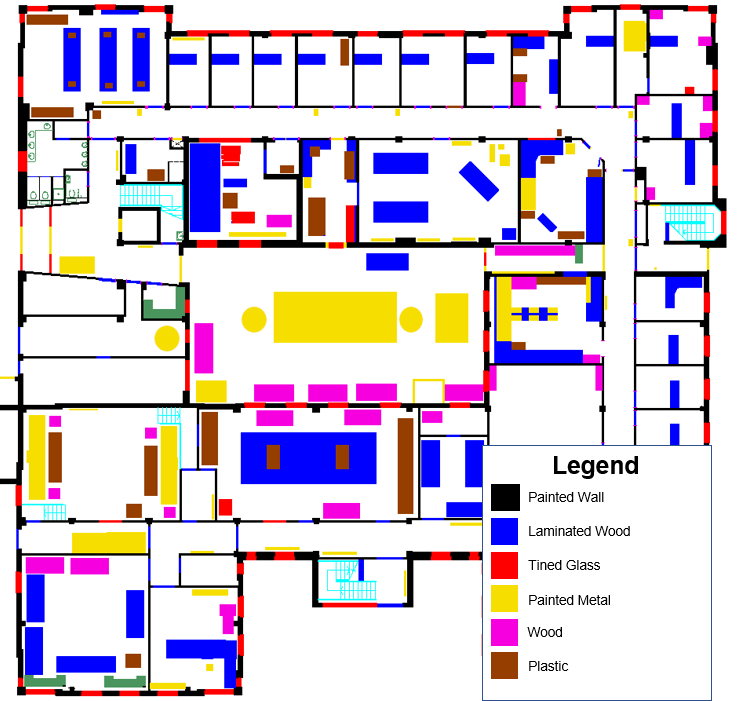}
	\caption{Material Map used in experimentation}
	\label{fi13:MMap}
	\vspace{-5mm}
\end{figure} 

\section{METHODOLOGY}
\begin{figure*}[ht]
\vspace{2mm}
	\centering
	\includegraphics[width=0.875\textwidth]{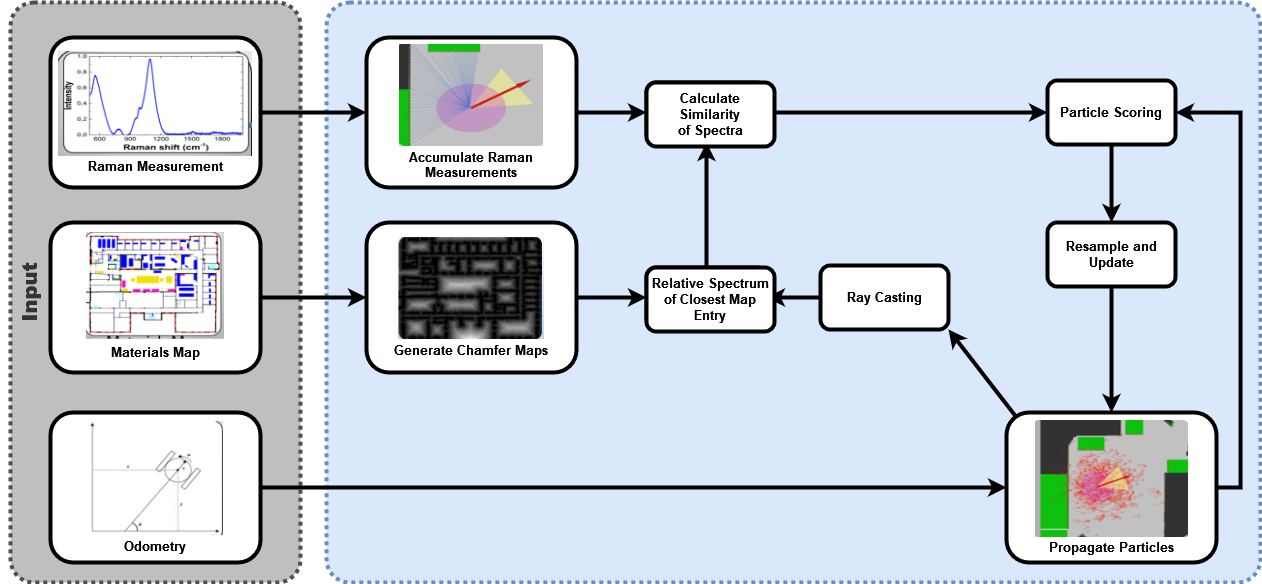}
	\caption{RaSpectLoc system diagram}
	\label{fi14:sysfig}
 	\vspace{-7mm}
\end{figure*}

Fig. 3 illustrates an overview of the RaSpectLoc system.
The system requires input odometry of the robot and a materials map with the Raman spectra embedded.
After initialisation, section IV-B details how Raman spectra are accumulated around the robot.
Section IV-C describes how the weight is calculated based on the spectral input from the mounted probe of the mobile robot.
The particles propagate according to the motion model of Eq. 2.
The system then repeats this process measuring the materials in the local environment around the robot with the Raman Probe and comparing them against the map.

\subsection{Material Floorplans}\label{AA}
The RaSpectLoc system offers a more innovative solution, utilizing spectral data obtained from a Raman probe. 
The system compares this data against previously recorded Raman spectra that are incorporated into the map.
In this work, each cell on the floorplan is associated with a unique Raman spectrum measurement. 
This map is used in the ray casting operation to determine the likelihood between the data and ends of the rays.
The materials represented in our floorplan for this work include various spectra for painted walls, laminate wood, painted metals, wood and plastic. 
These materials are broadly grouped based on similarity for visualization purposes, depicted in Fig.~\ref{fi13:MMap}.

The map is converted to an occupancy grid when initially uploaded to a map server.
If $\mathbb{M}$ is a set of possible 2D positions, the map can then be defined as
        $\mathbb{V} = \{ v_m ; m \in \mathbb{M} \}$. 
Each embedded spectrum is formally described by $\mathbb{I}_i = \{i_0, \dots, i_n\}$ where $i$ reflects the intensity of a particular wavelength.

	\vspace{-1mm}
\subsection{Raman Sensor}

\begin{figure}[ht]
	\centering
	\includegraphics[width=\columnwidth]{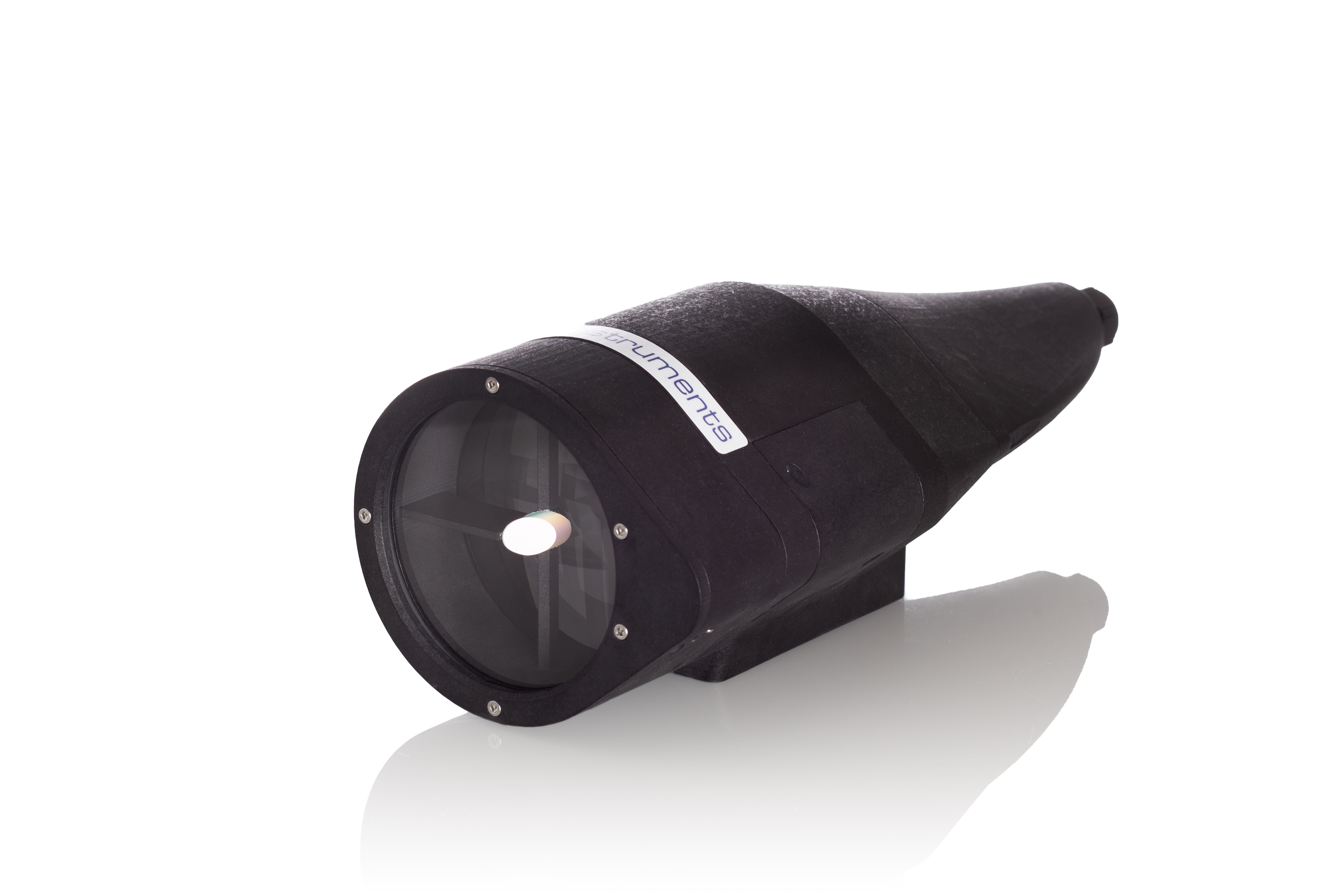}
	\caption{RP1000 Raman Probe~\cite{b38}}
	\label{fi31:spectrometer}
	\vspace{-4mm}
\end{figure} 

At the time $t$, the sensor on the robot generates a message consisting of a tuple of ranges, bearings, and Raman spectra, $\mathbb{Z}_t$.
Modern Raman spectra allow a non-invasive technique with background noise scattering removal for high-accuracy readings with little post-processing for smoothed spectral results.
RaSpectLoc can also operate without an explicit range sensor (i.e. only using bearing and spectra tuples) as explained in section IV-D.
In this case, raycasting is used as with the ``beam'' sensor model, and the range component of the likelihood is omitted.
The measurements are configured as $\mathbb{Z}_t = [\langle r_t^{k}, \Theta_t^{k}, \mathbb{I}_t^{k} \rangle; k = 1 \dots k]$, which represents the range from the depth camera, the bearing along the scan line, and the Raman spectrum respectively.
The scan line is assumed to be parallel to the ground plane and aligned with the horizontal axis of the mobile robot's sensor mounting point.
Raman measurements are taken by rotating the joint around the robot to complete a single scan message with $k$ measurements.
In the gazebo simulator, our gazebo plug-in can be mounted anywhere on a modelled robot and produce a rotating scan line. 

Our approach in RaSpectLoc aims to tackle the challenges faced in localisation for modular robots, which are currently dominated by geometry primitive recognition methods that require higher-end hardware.
We leverage the similarity between spectra that are read from the environment after normalisation and baseline correction is performed on them.
The information we gather is converted into chamfer maps, enabling the RaSpectLoc system to calculate the similarity for each particle.
Despite the advanced technology, current Raman probes still have a level of noise that must be accounted for when computing the similarity between observation and map. 
The shot, readout, thermal background, and baseline noise are Poisson distributed and assumed to be proportional to the square root of the number of photons detected.
Most Raman systems require proximity to the target in question. The IS-Instruments RP1000 probe, coupled to a HES2000, provides the capability of making Raman measurements at ranges over 1m. Such a probe is ideal for robot deployment and an ideal candidate for this application.

The noise in low response areas of spectra substantially affects the results using many comparison functions. 
When there is a significant number of (photonic) events, Poisson noise is indistinguishable from Gaussian noise as described by Larkin~\cite{b34} and also by Lewis et al.~\cite{b35}.
Therefore, we use a Squared exponential mapping function to convert spectrum distance to a likelihood, $P_s$, mapped between $0~\overrightarrow{}~1$:
\begin{equation}
    P_s\left( \mathbb{Z}_{t} | s_{t}, \mathbb{V}  \right) = \mathrm{exp} \left( \dfrac{-f\left( \mathbb{I}^k_t, \mathbb{I}^m \right)^{2}}{K} \right)
\end{equation}
where $f\left( \mathbb{I}^k_t, \mathbb{I}^m \right)$ is the distance function of the spectra $\mathbb{I}^k_t$ and $\mathbb{I}^m$ which is the spectra at endpoint of a raycasted beam $m$. 
While $K$ is a constant used to scale the different distance metrics into the same range.
Our approach supports previous state-of-the-art mathematical similarity functions for spectra~\cite{b20}, in addition to originally proposed methods.
	\vspace{-5mm}
\subsection{Spectral Similarity Functions} 
In RaSpectLoc, various similarity functions can be used to compare Raman spectra and calculate the particle weights. 
We will describe these functions in detail and their advantages and disadvantages.
The similarity functions described in this section are the Spectral Linear Kernel, Modified Euclidean distance~\cite{b20}, Wasserstein distance~\cite{b23}, Kullback-Leibler distance and Spectral Angle Mapping (SAM).
\subsubsection{Spectral Linear Kernel}
The spectral linear kernel (SLK), proposed by Conroy et al.~\cite{b23}, was designed for comparing Raman spectra. 
It can be expressed by:
\begin{equation}
    f_{slk}(\mathbb{I}^k_t, \mathbb{I}^m) = \! \sum_{i_n^k \in \mathbb{I}^k_t} \left( i^k_n \cdot i^m_n + \! \sum_{j = n - W}^{j = n + W}(i^k_n - i^k_j)(i^m_n - i^m_j)\! \right)
\end{equation}
where $\mathbb{I}^k$ and $\mathbb{I}^m$ are the input spectra.
The kernel considers the original intensity values at each wave number and includes the difference between the intensity of its neighbouring points on the spectrum.
The windowed difference product helps compare the relative shapes of the two spectra. 
However, relative to other functions mentioned in this paper it is slow and results can be poor when calculating the similarity over long windows of non-similar regions in the two spectra.

\subsubsection{Modified Euclidian Spectral Similarity metric (Mod. L2)}
The approach introduced by Khan et al.~\cite{b20} is an alternative to the Euclidian similarity calculated between each intensity.
This approach gives equal importance to the peaks of each spectrum but also rewards spectra that have shorter distances between the peaks.
The squared distance between intensities of the spectra can be calculated via $D^{i}_{\mathbb{I}}=(\mathbb{I}^k_i - \mathbb{I}^m_i)^2$. 
This can then be used under the conditions in equation (7) to calculate the Mod. L2 distance:
\begin{equation}
        \! f_{\! me}(\mathbb{I}^k_t,\mathbb{I}^m)^2 = 
        \begin{cases}
            \! \displaystyle \sum_{i_n^k \in \mathbb{I}^k_t} \! \frac{1}{w} \ast D^{i}_{\mathbb{I}},& \! \text{if } x_i \neq 0 \text{ AND } z_i > x_i \\
            \! \displaystyle \sum_{i_n^k \in \mathbb{I}^k_t} \! w \ast D^{i}_{\mathbb{I}},& \text{if } x_i = 0 \text{ AND }  z_i > x_i \\
            \! \displaystyle \sum_{i_n^k \in \mathbb{I}^k_t} \! D^{i}_{\mathbb{I}}, \!             & \text{otherwise}
        \end{cases}
\end{equation}
The function above calculates a distance between the spectra using a weight, $w$, found by:
\begin{equation}
 w = \frac{max(\mathbb{I}^k_t)}{1-max(\mathbb{I}^k_t)} 
\end{equation}
The weight $w$ is utilized as a penalty/reward scheme for relative changes in individual intensities at each wavelength.
The drawback of this function is that the weight does not work when the $maxQ$ value is less than 0.5. 
To resolve this issue, the weight is inverted to correct the error from the original equation.

\subsubsection{Wassertein Distance}
The Wasserstein distance, also known as the Earth Mover's Distance, is a metric that quantifies the effort required to transform one distribution into another.
It was suggested in previous literature to be an effective method of evaluating the similarity of spectra by Gao et al.~\cite{b35}. 
Despite its accuracy, the Wasserstein distance is complex and slow compared to the other functions mentioned in this section. 
This is explored in Section V.
\begin{equation}
    f_W(\mathbb{I}^k_t,\mathbb{I}^m) = min \{ E(t) | t : \mathbb{I}^k_t \rightarrow  \mathbb{I}^m \}
\end{equation}
Where $t$ is a transportation plan that maps the elements of $\mathbb{I}^k_t$ to the elements of $\mathbb{I}^m$, and $E(t)$ 
    is the cost of the transportation plan $t$, which is typically represented as the sum of the distances between the corresponding vector elements in $\mathbb{I}^k_t$ and $\mathbb{I}^m$.

\subsubsection{Kullback-Leibler Divergence}
The KL-Divergence is a highly efficient yet effective distance to compare two equally binned distributions. 
Formally it is a generalisation on the L2 Norm and is defined as the 
\begin{equation}
    f_{kl}(\mathbb{I}^k_t||\mathbb{I}^m) = \sum_{i_n^k \in \mathbb{I}^k_t} i^k_n \log \frac{i^k_n}{i^m_n}
\end{equation}
The KL-divergence quantifies the information lost when approximating $\mathbb{I}^k_t$ with $\mathbb{I}^m$. 
It is a non-symmetric metric, meaning that $f_{KL}(\mathbb{I}^k_t || \mathbb{I}^m) \neq f_{KL}(\mathbb{I}^k_t || \mathbb{I}^m)$.
The KL divergence is sensitive to tiny differences between distributions, which is desirable in the context of Raman spectra, where little variations in intensity values can indicate the presence of 
    different materials.

\subsubsection{Spectral Angle Mapping}
The Spectral Angle Mapping (SAM) function is a similarity metric commonly used to calculate the angle between two spectra in a high-dimensional space.
Each dimension represents a different Raman shift. 
The angle between the two spectra shows their similarity, with a smaller angle indicating higher similarity. 
Formally the function can be expressed by:
\begin{equation}
f_{sam}(\mathbb{I}^k_t||\mathbb{I}^m) = \cos^{-1} \left(\frac{\mathbb{I}^k_t \cdot \mathbb{I}^m}{\left\lVert \mathbb{I}^k_t\right\rVert \left\lVert \mathbb{I}^m \right\rVert}\right)
\end{equation}
SAM is founded on the idea that similar spectra should have similar shapes and relative intensities across different Raman shifts benefiting from the similarity angle returned around the shape of each peak.
Furthermore, the function is also scale-invariant, meaning it is not affected by differences in the absolute intensity of the spectra.
However, SAM is quite sensitive to noise and variations of the baseline of the Raman spectra since they affect the shape of the peaks.
SAM also fails to consider the intensity of the spectra and assumes spectra are linearly mixed.
This assumption may not be the case for some materials in the environment.

\subsection{Sensor Models}
We adapt both the ``beam'' and ``likelihood-field'' sensor models for use with Raman Spectra.

The Beam model depends on ranges as part of the weight for each particle.
This is because the beam model depends on the depth of the scan from the environment and the end of the beam projected into the map, $\mathbb{V}$.
While the probability of an observation given the map and pose for the beam model can be expressed by a weighted sum of the two probabilities:
\begin{equation}
\begin{split}
    P_r(\mathbb{Z}_{t}^{k} | s_{t}^{i}, \mathbb{V}) = &\ 
       \epsilon_{R} P_{R}(\mathbb{Z}_{t}^{k} | s_{t}^{i}, \mathbb{V}) +
    \epsilon_{m} P_{m}(\mathbb{Z}_{t}^{k} | s_{t}^{i}, \mathbb{V}) 
	\label{eq:range_mat_Pr}
\end{split}
\end{equation}
For the Beam-model of RaSpectLoc, the likelihood of a particle is defined by two user-defined weights, $\epsilon_R$ and $\epsilon_m$. 
When $\epsilon_m$ is zero, the likelihood is equivalent to the standard RMCL. 
The optimal weight configuration is evaluated in Section V. 
Unlike range scanners, the standard deviation $\sigma^{'}$ of equation 2 cannot be linked to the sensor's physical properties. 
Instead, it is estimated from the prior of each material spectra on the map, avoiding the need for tuning.
However, the Beam Model has drawbacks to its approach as it lacks smoothness since it depends on the resolution of the map.
This is worse for mobile robots as it becomes more memory dependent to increase the number of beams and resolution of the map.

The likelihood-field model provides a smoother approach providing gradients between each cell.
Furthermore, it can use ranges, bearing and material likelihood or only bearing and material likelihoods.
The likelihood field model in our approach generates a chamfer map to the nearest occupied cell for each material based on equation~\ref{eq:2}:
\begin{equation}
    C_{i,j}\left(\mathbb{I}\right) = \min_{k,l}\left(C_{k,l} \left( \mathbb{I} \right) + f \left( \mathbb{I}_{i,j}, \mathbb{I}_{k,l} \right) \right)
\end{equation}
Where $f$ is the chosen spectral likelihood function mentioned in Section III-B.
For every $ \mathbb{Z}_t = [\langle \Theta_t^{k}, \mathbb{I}_t^{k} \rangle; k = 1 \dots k]$, the raycasting is performed. 
Instead of comparing the range with the map, the spectral likelihood can be used to calculate the cost:
\begin{equation}
    P_r(\mathbb{Z}_{t}^{k} | s_{t}^{i}, \mathbb{V}) = P_{m}(\mathbb{Z}_{t}^{k} | s_{t}^{i}, \mathbb{V})
\end{equation}
This method is a combination of the beam and likelihood model. 
Hence, in Eq.~\ref{eq:range_mat_Pr}, when $\epsilon_R$ is zero, the approach solely relies on the Raman spectra information in the floorplan and material similarity likelihood. 

The nearest occupied cell is identified through similar AMCL range-based raycasting. 
The distances obtained are utilized to improve the smoothness of Eq.~\ref{eq:range_mat_Pr}, which suggests that the likelihood of the spectra similarity is proportional to the distribution of material similarity in terms of angle. 
As a result, this approach is scale-invariant, provided that the aspect ratio of the map is maintained.

\section{Experiments and Results}
Our research focuses on MCL-based localisation using Raman Spectra from a single probe. 
This section presents the overall performance improvements of our system over other state-of-the-art localisation approaches. 
This dataset will be released to the community to support future research. 
RaSpectLoc can choose between the beam or the likelihood field model which are described in Section IV-D. 
The ablation and parameter exploration section, V-B, illustrates the difference in the performance of RaSpectLoc due to changes in likelihood functions and configurations of $\epsilon$. 
Whereas the Quantitative Results section V-C will compare RaSpectLoc with various state-of-the-art baselines.
These include, AMCL~\cite{b28}, SeDAR~\cite{b1}, visual pose recognition (PoseNet~\cite{b32}), range-based SLAM (GMapping~\cite{b33}) and Monocular SLAM (ORBSLAM 3~\cite{b31}).
	\vspace{-2mm}

\begin{table}[htp]
	\centering
	\resizebox{\columnwidth}{!}{%
	\begin{tabular}{|ccccccc|}
	\hline
	\multicolumn{7}{|c|}{Spectral Similarity Function Analysis (m)} \\ \hline
	\multicolumn{1}{|c|}{Function} &
	  \multicolumn{1}{c|}{RMSE} &
	  \multicolumn{1}{c|}{Mean} &
	  \multicolumn{1}{c|}{Median} &
	  \multicolumn{1}{c|}{SD} &
	  \multicolumn{1}{c|}{Min} &
	  Max \\ \hline
	\multicolumn{1}{|c|}{SLK} &
	  \multicolumn{1}{c|}{0.18} &
	  \multicolumn{1}{c|}{0.17} &
	  \multicolumn{1}{c|}{0.16} &
	  \multicolumn{1}{c|}{0.07} &
	  \multicolumn{1}{c|}{0.03} &
	  0.38 \\ \hline
	\multicolumn{1}{|c|}{Mod.L2} &
	  \multicolumn{1}{c|}{\textbf{0.14}} &
	  \multicolumn{1}{c|}{\textbf{0.13}} &
	  \multicolumn{1}{c|}{0.12} &
	  \multicolumn{1}{c|}{\textbf{0.06}} &
	  \multicolumn{1}{c|}{0.03} &
	  0.41 \\ \hline
	\multicolumn{1}{|c|}{Wasserstein} &
	  \multicolumn{1}{c|}{0.17} &
	  \multicolumn{1}{c|}{0.14} &
	  \multicolumn{1}{c|}{0.13} &
	  \multicolumn{1}{c|}{0.09} &
	  \multicolumn{1}{c|}{0.02} &
	  0.44 \\ \hline
	\multicolumn{1}{|c|}{KL.D} &
	  \multicolumn{1}{c|}{0.15} &
	  \multicolumn{1}{c|}{0.13} &
	  \multicolumn{1}{c|}{\textbf{0.11}} &
	  \multicolumn{1}{c|}{0.08} &
	  \multicolumn{1}{c|}{\textbf{0.02}} &
	  \textbf{0.34} \\ \hline
	\multicolumn{1}{|c|}{SAM} &
	  \multicolumn{1}{c|}{0.18} &
	  \multicolumn{1}{c|}{0.16} &
	  \multicolumn{1}{c|}{0.14} &
	  \multicolumn{1}{c|}{0.09} &
	  \multicolumn{1}{c|}{0.03} &
	  0.42 \\ \hline
	\end{tabular}%
	}
	\caption{Spectral Similarity Function Experiments}
	\label{tab:Spec-analATE}
	\vspace{-7mm}
	\end{table}

\begin{table*}[t]
    \vspace{4mm}
	\centering
	\resizebox{\textwidth}{!}{%
	\begin{tabular}{|ccccccccccc|}
	\hline
	\multicolumn{11}{|c|}{Spectral Similarity Function Analysis (RPE)} \\ \hline
	\multicolumn{1}{|c|}{Function} &
	  \multicolumn{1}{c|}{\begin{tabular}[c]{@{}c@{}}Translational\\ RMSE(m)\end{tabular}} &
	  \multicolumn{1}{c|}{\begin{tabular}[c]{@{}c@{}}Rotational\\ RMSE(deg)\end{tabular}} &
	  \multicolumn{1}{c|}{\begin{tabular}[c]{@{}c@{}}Translational\\ Mean(m)\end{tabular}} &
	  \multicolumn{1}{c|}{\begin{tabular}[c]{@{}c@{}}Rotational\\ Mean(deg)\end{tabular}} &
	  \multicolumn{1}{c|}{\begin{tabular}[c]{@{}c@{}}Translational\\ Median(m)\end{tabular}} &
	  \multicolumn{1}{c|}{\begin{tabular}[c]{@{}c@{}}Rotational\\ Median(deg)\end{tabular}} &
	  \multicolumn{1}{c|}{\begin{tabular}[c]{@{}c@{}}Translational\\ SD(m)\end{tabular}} &
	  \multicolumn{1}{c|}{\begin{tabular}[c]{@{}c@{}}Rotational\\ SD(deg)\end{tabular}} &
	  \multicolumn{1}{c|}{\begin{tabular}[c]{@{}c@{}}Translational\\ Max(m)\end{tabular}} &
	  \begin{tabular}[c]{@{}c@{}}Rotational\\ Max(deg)\end{tabular} \\ \hline
	\multicolumn{1}{|c|}{Wasserstein} &
	  \multicolumn{1}{c|}{0.06} &
	  \multicolumn{1}{c|}{\textbf{2.02}} &
	  \multicolumn{1}{c|}{0.05} &
	  \multicolumn{1}{c|}{1.39} &
	  \multicolumn{1}{c|}{0.04} &
	  \multicolumn{1}{c|}{0.91} &
	  \multicolumn{1}{c|}{0.03} &
	  \multicolumn{1}{c|}{\textbf{1.46}} &
	  \multicolumn{1}{c|}{0.22} &
	  \textbf{9.67} \\ \hline
	\multicolumn{1}{|c|}{SLK} &
	  \multicolumn{1}{c|}{0.07} &
	  \multicolumn{1}{c|}{2.32} &
	  \multicolumn{1}{c|}{0.06} &
	  \multicolumn{1}{c|}{1.58} &
	  \multicolumn{1}{c|}{0.05} &
	  \multicolumn{1}{c|}{0.97} &
	  \multicolumn{1}{c|}{0.04} &
	  \multicolumn{1}{c|}{1.70} &
	  \multicolumn{1}{c|}{0.20} &
	  11.9 \\ \hline
	\multicolumn{1}{|c|}{KL.D} &
	  \multicolumn{1}{c|}{\textbf{0.05}} &
	  \multicolumn{1}{c|}{2.03} &
	  \multicolumn{1}{c|}{\textbf{0.04}} &
	  \multicolumn{1}{c|}{\textbf{1.34}} &
	  \multicolumn{1}{c|}{0.04} &
	  \multicolumn{1}{c|}{\textbf{0.81}} &
	  \multicolumn{1}{c|}{\textbf{0.02}} &
	  \multicolumn{1}{c|}{1.52} &
	  \multicolumn{1}{c|}{\textbf{0.16}} &
	  10.86 \\ \hline
	\multicolumn{1}{|c|}{Mod.L2} &
	  \multicolumn{1}{c|}{0.06} &
	  \multicolumn{1}{c|}{2.23} &
	  \multicolumn{1}{c|}{0.05} &
	  \multicolumn{1}{c|}{1.5} &
	  \multicolumn{1}{c|}{\textbf{0.04}} &
	  \multicolumn{1}{c|}{0.95} &
	  \multicolumn{1}{c|}{0.03} &
	  \multicolumn{1}{c|}{1.64} &
	  \multicolumn{1}{c|}{0.17} &
	  10.17 \\ \hline
	\multicolumn{1}{|c|}{SAM} &
	  \multicolumn{1}{c|}{0.07} &
	  \multicolumn{1}{c|}{2.34} &
	  \multicolumn{1}{c|}{0.06} &
	  \multicolumn{1}{c|}{1.57} &
	  \multicolumn{1}{c|}{0.05} &
	  \multicolumn{1}{c|}{0.96} &
	  \multicolumn{1}{c|}{0.04} &
	  \multicolumn{1}{c|}{1.73} &
	  \multicolumn{1}{c|}{0.31} &
	  9.94 \\ \hline
	\end{tabular}%
	}
	\caption{Spectral Similarity Function Analysis (RPE)}
	\label{tab:RPE-set}
	\vspace{-7mm}
	\end{table*}  
 
\subsection{Experimental Simulation Setup}
% \vspace{-1mm}
The following sections evaluate the performance of RaSpectLoc using a dataset consisting of 5 mobile robot trajectories around a University. 
The dataset includes Raman Spectra, RGBD images and odometry information. 
Additionally, a material-embedded floorplan with Raman spectra at each coordinate is provided.

To quantitatively evaluate the performance of RaSpectLoc and state-of-the-art methods, the Absolute Trajectory Error (ATE) and the Relative Pose Error (RPE) metrics presented by Strum et al.~\cite{b27} are used. 
The ATE calculates the rigid transformation estimation between the two paths.
The RPE evaluates the relative rotation and relative translation for all points in the path, ignoring the effects of drift over time in the trajectory. 
The RMSE, mean, median, standard deviation, minimum distance and maximum distance values of the residual position error are provided for the resulting ATE and RPE of the experiments.
% We also aimed to observe the time taken to converge from a global initialisation at run time (TTC).
% The TTC is the time taken since run-time for the localisation to converge within 5cm of the ground truth. 
% Usually, in global initialisation experiments, only the ATE is considered however this metric can be undescriptive.
For evaluation in Table~\ref{tab:Spec-analATE}, AMCL, SeDAR~\cite{b1} and RaSpectLoc is given a coarse initialisation with standard deviations of 2.0m in (x, y) and 2.0 radians in $\theta$. 
The system was run with a maximum of 1000 particles placed around the covariance ellipse. The error was recorded for each new set of accumulated Raman spectra into the published tuple $\mathbb{Z}_t$.  

\subsection{Ablation and parameter exploration}
\vspace{-1mm}
\begin{figure}[h]
	\centering
	\includegraphics[width=0.9\columnwidth, height=5.0cm]{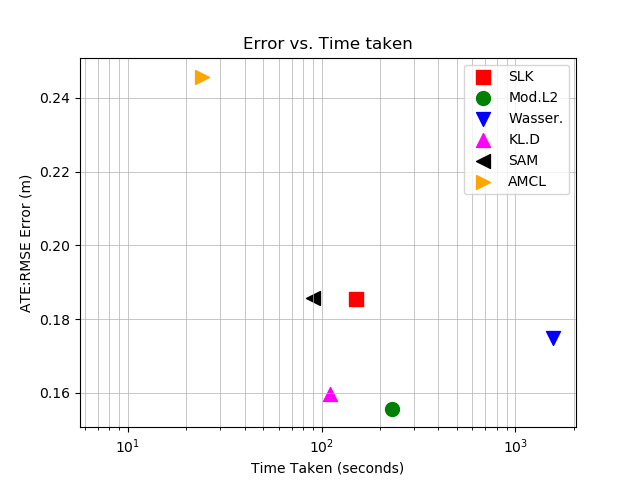}
	\caption{ATE:RMSE error against time taken to complete the trajectory}
	\label{ATE_TIME_PLOT}
	\vspace{-2mm}
\end{figure} 
This section focuses on the performance of the RaSpectLoc system using various spectral similarity functions and optimal weights, $\epsilon$, to accurately compare against modern localisation approaches. 
We evaluate the performance using the ATE and RPE of the produced trajectories with the likelihood-field model. 
Additionally, we consider the accuracy of determining the correct spectrum after adding Raman-probe shot noise. 
This approach should provide a more realistic evaluation of the similarity functions compared to post-processed spectra that do not contain these intensity artefacts or noise.
For Eq. 12 the value of $\beta$ is set to 1.
% \vspace{-2mm}
The results in Tables~\ref{tab:Spec-analATE} and~\ref{tab:RPE-set} indicate that the Mod. L2 function produces the best accuracy and performance for RaSpectLoc. 
The results of the Mod. L2 function is because it provides a descriptive way of examining the similarity between peaks with a reward and punishment scheme. 
In testing, the Wasserstein distance takes significantly longer to process than other functions, often leading to performance issues during periods of high message frequency.
This is highlighted in Fig.~\ref{ATE_TIME_PLOT} by the logarithmic axis for the time taken to complete a 5min 55sec rosbag recording. One obvious limitation of RaSpectLoc from this graph is the increased run-time compared to AMCL, although this comes with a significant reduction in error.
	
\begin{table}[t]
	\centering
	\resizebox{\columnwidth}{!}{%
	\begin{tabular}{|cccccccc|}
	\hline
	\multicolumn{8}{|c|}{Weight configuration tests (ATE)} \\ \hline
	\multicolumn{2}{|c|}{Weights} &
	  \multicolumn{1}{c|}{\multirow{2}{*}{RMSE}} &
	  \multicolumn{1}{c|}{\multirow{2}{*}{Mean}} &
	  \multicolumn{1}{c|}{\multirow{2}{*}{Median}} &
	  \multicolumn{1}{c|}{\multirow{2}{*}{SD}} &
	  \multicolumn{1}{c|}{\multirow{2}{*}{Min}} &
	  \multirow{2}{*}{Max} \\ \cline{1-2}
	\multicolumn{1}{|c|}{$\epsilon_R$} &
	  \multicolumn{1}{c|}{$\epsilon_m$} &
	  \multicolumn{1}{c|}{} &
	  \multicolumn{1}{c|}{} &
	  \multicolumn{1}{c|}{} &
	  \multicolumn{1}{c|}{} &
	  \multicolumn{1}{c|}{} &
	   \\ \hline
	\multicolumn{1}{|c|}{0.2} &
	  \multicolumn{1}{c|}{0.8} &
	  \multicolumn{1}{c|}{0.19} &
	  \multicolumn{1}{c|}{0.17} &
	  \multicolumn{1}{c|}{0.15} &
	  \multicolumn{1}{c|}{\textbf{0.09}} &
	  \multicolumn{1}{c|}{\textbf{0.01}} &
	  0.43 \\ \hline
	\multicolumn{1}{|c|}{0.4} &
	  \multicolumn{1}{c|}{0.6} &
	  \multicolumn{1}{c|}{0.19} &
	  \multicolumn{1}{c|}{0.17} &
	  \multicolumn{1}{c|}{0.16} &
	  \multicolumn{1}{c|}{0.09} &
	  \multicolumn{1}{c|}{0.04} &
	  \textbf{0.38} \\ \hline
	\multicolumn{1}{|c|}{0.5} &
	  \multicolumn{1}{c|}{0.5} &
	  \multicolumn{1}{c|}{\textbf{0.18}} &
	  \multicolumn{1}{c|}{\textbf{0.15}} &
	  \multicolumn{1}{c|}{\textbf{0.13}} &
	  \multicolumn{1}{c|}{0.10} &
	  \multicolumn{1}{c|}{0.03} &
	  0.50 \\ \hline
	\multicolumn{1}{|c|}{0.6} &
	  \multicolumn{1}{c|}{0.4} &
	  \multicolumn{1}{c|}{0.22} &
	  \multicolumn{1}{c|}{0.20} &
	  \multicolumn{1}{c|}{0.20} &
	  \multicolumn{1}{c|}{0.10} &
	  \multicolumn{1}{c|}{0.08} &
	  0.47 \\ \hline
	\multicolumn{1}{|c|}{0.8} &
	  \multicolumn{1}{c|}{0.2} &
	  \multicolumn{1}{c|}{0.21} &
	  \multicolumn{1}{c|}{0.19} &
	  \multicolumn{1}{c|}{0.19} &
	  \multicolumn{1}{c|}{0.11} &
	  \multicolumn{1}{c|}{0.08} &
	  0.45 \\ \hline
	\end{tabular}%
	}
	\caption{Weight experiments ablation tests}
	\label{tab:weight-tests}
	\vspace{-7mm}
\end{table}

As shown in Table~\ref{tab:weight-tests}, the performance of RaSpectLoc varies with the weight distribution.
The best-performing weights are 0.5 for the material and ranges. 

Compared to Table~\ref{tab:Spec-analATE} the results still show that a `spectra-only' approach outweighs the combined range and spectral data approach seen in Table~\ref{tab:weight-tests}.
This result is likely due to geometric cues in a hallway environment.
These are relatively weak compared to the fine-grained spectral data that provide a greater understanding of the current pose for the system.
Findings in these experiments are highlighted in Section V-C, where we will compare RaSpectLoc's performance against other state-of-the-art systems.
	\vspace{-1mm}
\subsection{Quantitative Results}
Tables~\ref{tab:ATE-comparisons} and~\ref{tab:RPE-comparisons} showcase the results of our tests against various state-of-the-art localisation approaches in terms of Absolute Trajectory Error (ATE) and Relative Pose Error (RPE) metrics. 
Experiments are compared with $\epsilon_R$ and $\epsilon_m$ set to 0.5 and 0.5, respectively, as per the optimal configuration found in Section V-B for our combined localisation model with the Mod. L2 likelihood function (Section IV-C2). 
These tests are performed using the same RGB images and depth scan messages along a single trajectory. 
This separation allows us to compare our results against three non-MCL approaches: ORBSLAM, GMapping and PoseNet.

% \usepackage{graphicx}
\begin{table}[thbp]
	\centering
	\resizebox{\columnwidth}{!}{%
	\begin{tabular}{|ccccccc|}
	\hline
	\multicolumn{7}{|c|}{Average Trajectory Error (m)} \\ \hline
	\multicolumn{1}{|c|}{Approach} &
	  \multicolumn{1}{c|}{RMSE} &
	  \multicolumn{1}{c|}{Mean} &
	  \multicolumn{1}{c|}{Median} &
	  \multicolumn{1}{c|}{SD} &
	  \multicolumn{1}{c|}{Min} &
	  Max \\ \hline
	\multicolumn{1}{|c|}{AMCL} &
	  \multicolumn{1}{c|}{0.24} &
	  \multicolumn{1}{c|}{0.21} &
	  \multicolumn{1}{c|}{0.2} &
	  \multicolumn{1}{c|}{0.11} &
	  \multicolumn{1}{c|}{0.04} &
	  0.95 \\ \hline
	\multicolumn{1}{|c|}{SeDAR} &
	  \multicolumn{1}{c|}{0.19} &
	  \multicolumn{1}{c|}{0.16} &
	  \multicolumn{1}{c|}{0.14} &
	  \multicolumn{1}{c|}{0.10} &
	  \multicolumn{1}{c|}{\textbf{0.02}} &
	  0.55 \\ \hline
	\multicolumn{1}{|c|}{ORBSLAM3 (MONO)} &
	  \multicolumn{1}{c|}{7.17} &
	  \multicolumn{1}{c|}{6.54} &
	  \multicolumn{1}{c|}{6.81} &
	  \multicolumn{1}{c|}{2.93} &
	  \multicolumn{1}{c|}{0.55} &
	  11.57 \\ \hline
	\multicolumn{1}{|c|}{GMapping} &
	  \multicolumn{1}{c|}{0.71} &
	  \multicolumn{1}{c|}{0.63} &
	  \multicolumn{1}{c|}{0.57} &
	  \multicolumn{1}{c|}{0.31} &
	  \multicolumn{1}{c|}{0.32} &
	  1.43 \\ \hline
	\multicolumn{1}{|c|}{Posenet} &
	  \multicolumn{1}{c|}{4.64} &
	  \multicolumn{1}{c|}{2.58} &
	  \multicolumn{1}{c|}{1.42} &
	  \multicolumn{1}{c|}{3.85} &
	  \multicolumn{1}{c|}{0.02} &
	  25.66 \\ \hline
	\multicolumn{1}{|c|}{RaSpectLoc (combined)} &
	  \multicolumn{1}{c|}{0.18} &
	  \multicolumn{1}{c|}{0.15} &
	  \multicolumn{1}{c|}{0.13} &
	  \multicolumn{1}{c|}{0.10} &
	  \multicolumn{1}{c|}{0.03} &
	  0.50 \\ \hline
	\multicolumn{1}{|c|}{RaSpectLoc (Materials only)} &
	  \multicolumn{1}{c|}{\textbf{0.14}} &
	  \multicolumn{1}{c|}{\textbf{0.13}} &
	  \multicolumn{1}{c|}{\textbf{0.12}} &
	  \multicolumn{1}{c|}{\textbf{0.06}} &
	  \multicolumn{1}{c|}{0.03} &
	  \textbf{0.41} \\ \hline
	\end{tabular}%
	}
	\caption{ATE: Baseline localisation comparisons}
	\label{tab:ATE-comparisons}
	\vspace{-7mm}
	\end{table}

% \usepackage{graphicx}
\begin{table}[thbp]
	\centering
	\resizebox{\columnwidth}{!}{%
	\begin{tabular}{|cccccc|}
	\hline
	\multicolumn{6}{|c|}{Relative Pose Error (m/deg)} \\ \hline
	\multicolumn{1}{|c|}{Approach} &
	  \multicolumn{1}{c|}{RMSE} &
	  \multicolumn{1}{c|}{Mean} &
	  \multicolumn{1}{c|}{Median} &
	  \multicolumn{1}{c|}{SD} &
	  Max \\ \hline
	\multicolumn{1}{|c|}{AMCL} &
	  \multicolumn{1}{c|}{0.07/2.68} &
	  \multicolumn{1}{c|}{0.05/1.68} &
	  \multicolumn{1}{c|}{0.04/0.68} &
	  \multicolumn{1}{c|}{0.05/2.09} &
	  0.37/11.89 \\ \hline
	\multicolumn{1}{|c|}{SeDAR} &
	  \multicolumn{1}{c|}{0.06/2.21} &
	  \multicolumn{1}{c|}{0.04/1.5} &
	  \multicolumn{1}{c|}{0.04/0.9} &
	  \multicolumn{1}{c|}{0.03/1.72} &
	  0.2/12.47 \\ \hline
	\multicolumn{1}{|c|}{ORBSLAM3 (MONO)} &
	  \multicolumn{1}{c|}{0.42/14.7} &
	  \multicolumn{1}{c|}{0.41/5.78} &
	  \multicolumn{1}{c|}{0.41/0.86} &
	  \multicolumn{1}{c|}{0.10/13.58} &
	  0.77/61.60 \\ \hline
	\multicolumn{1}{|c|}{GMapping} &
	  \multicolumn{1}{c|}{0.40/26.65} &
	  \multicolumn{1}{c|}{0.18/13.05} &
	  \multicolumn{1}{c|}{0.31/8.5} &
	  \multicolumn{1}{c|}{0.35/23.23} &
	  1.49/92.65 \\ \hline
	\multicolumn{1}{|c|}{Posenet} &
	  \multicolumn{1}{c|}{4.72/47.70} &
	  \multicolumn{1}{c|}{2.23/28.40} &
	  \multicolumn{1}{c|}{0.89/9.8} &
	  \multicolumn{1}{c|}{4.16/38.33} &
	  28.96/179.6 \\ \hline
	\multicolumn{1}{|c|}{RaSpectLoc (combined)} &
	  \multicolumn{1}{c|}{0.06/2.31} &
	  \multicolumn{1}{c|}{0.06/1.52} &
	  \multicolumn{1}{c|}{0.05/0.94} &
	  \multicolumn{1}{c|}{0.03/1.74} &
	  0.21/9.77 \\ \hline
	\multicolumn{1}{|c|}{RaSpectLoc (Materials)} &
	  \multicolumn{1}{c|}{\textbf{0.05/2.03}} &
	  \multicolumn{1}{c|}{\textbf{0.04/1.34}} &
	  \multicolumn{1}{c|}{\textbf{0.04/0.81}} &
	  \multicolumn{1}{c|}{\textbf{0.02/1.52}} &
	  \textbf{0.16/10.86} \\ \hline
	\end{tabular}%
	}
	\caption{RPE: Baseline localisation comparisons}
	\label{tab:RPE-comparisons}
		\vspace{-4mm}
\end{table}

Our experiments revealed that RaSpectLoc consistently outperforms the other approaches in accuracy, robustness, ATE and RPE metrics. 
The results showed that our method was able to accurately estimate the 3-DoF pose of the robot even in challenging environments with noise and dynamic obstacles. 
Furthermore, our experiments highlighted the ability of RaSpectLoc to scale to larger environments and handle noise-riddled Raman Spectra, making it a highly competitive solution for real-world applications. 

% \vspace{-2mm}
 
\section{Conclusion}
In conclusion, this paper has presented a novel approach for mobile robot localisation that harnesses the power of Raman spectra. 
With its ability to recognise subtle material-composition landmarks, RaSpectLoc can outperform traditional RGBD systems.
This recognition has significant implications for localisation in challenging environments, evident from our practical experimental section (V-C). 
The future of this field is even more promising, as we envision a future where robots can exploit spectral data to navigate and map hazardous environments, such as nuclear decommissioning sites. 
Such a SLAM system would provide a more detailed understanding of the environment while reducing the risk posed to human workers. 
Raman spectra could also assist in detecting anomalies in these high-risk locations with the potential to revolutionize the field and greatly benefit society.

\section{Acknowledgements}
This work was partially supported by Industrial 3D Robotics (I3D), IS-Instruments and partially funded by the EPSRC under grant agreement EP/S035761/1.

\vspace{12pt}

\end{document}